# KINETOSTATIC ANALYSIS AND SOLUTION CLASSIFICATION OF A CLASS OF PLANAR TENSEGRITY MECHANISMS

P. Wenger[1] and D. Chablat[1]

[1] Laboratoire des Sciences du Numérique de Nantes, UMR CNRS 6004

Ecole Centrale de Nantes, 1, Rue de la Noe, 44321 Nantes, France,

email: [philippe.wenger, damien.chablat]@ls2n.fr

**Abstract:** Tensegrity mechanisms are composed of rigid and tensile parts that are in equilibrium. They are interesting alternative designs for some applications, such as modelling musculo-skeleton systems. Tensegrity mechanisms are more difficult to analyze than classical mechanisms as the static equilibrium conditions that must be satisfied generally result in complex equations. A class of planar one-degree-of-freedom tensegrity mechanisms with three linear springs is analyzed in detail for the sake of systematic solution classifications. The kinetostatic equations are derived and solved under several loading and geometric conditions. It is shown that these mechanisms exhibit up to six equilibrium configurations, of which one or two are stable, depending on the geometric and loading conditions. Discriminant varieties and cylindrical algebraic decomposition combined with Groebner base elimination are used to classify solutions as a function of the geometric, loading and actuator input parameters.

## 1 INTRODUCTION

A tensegrity structure is an assembly of compressive elements (struts) and tensile elements (cable, springs) held together in equilibrium [1, 2, 3].Their inherent interesting features (low inertia, natural compliance and deployability) make them suitable in several applications. They can also be used as preliminary models in musculo-skeleton systems to analyze animal and human movements [4, 5]. A spine can be modelled by stacking a number of suitable tensegrity modules, see for example [6]. Figure 1 illustrates some configurations of a serial assembly of planar tensegrity mechanisms similar to the ones studied in this paper, suitable for a plane model of a spine or of an animal neck.

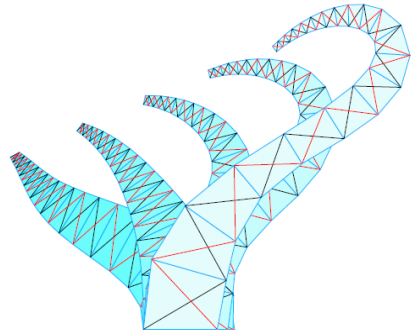

**Fig 1: A serial assembly of elementary tensegrity mechanisms mimicking an animal neck (illustration from [7])**

The frame of this work is a preliminary step of a large collaborative project with the Museum National d'Histoire Naturelle, called "AVINECK: the neck of birds, an arm for robots". This project aims at understanding the functional behavior of bird necks and using them as bio-inspired models for the design of innovative robotic arms. In this project, several bird neck models will be considered, starting with a planar model made of tensegrity mechanisms. A first step of the project is to seek for appropriate simple models that lend themselves to symbolic calculations in order to better understand their kinetostatic behavior and to ease further optimizations for design purposes.

A tensegrity mechanism is obtained when one or several elements are actuated. Most research efforts on tensegrity mechanisms have been conducted recently (see for example [6-10] and references therein). The input and output equations of a tensegrity mechanism are derived after solving the equilibrium conditions. These conditions are generally obtained by minimizing the potential energy. Symbolic derivation of minimization conditions often generates complex equations and inequations that are difficult to solve. Due to their relative simplicity, planar tensegrity mechanisms are more suitable for algebraic computations. A planar 2-DOF planar tensegrity mechanism was analyzed by Arsenault in [8], where kinetostatics and dynamics models as well as some simplified workspaces were studied. Recently, Boehler [10] proposed a more complete definition of the workspace of that 2-DOF mechanism as well as a higher-order continuation based method to evaluate it.

This paper goes beyond a first study presented in [11] and analyses in details a family of one-degree-of-freedom planar tensegrity mechanisms made of one base telescopic rod, two crossed fixed-length rods and three connecting springs (see fig. 1). This type of mechanism was studied in [12] in the particular case of symmetric geometric and loading conditions. The equilibrium configurations were solved for a set of geometric parameters and for one actuator input value. Here, these planar tensegrity mechanisms are analyzed in a more systematic way and in more detail, with the goal of understanding in depth the evolution of the number of stable and unstable solutions as a function of the geometric parameters, the loading conditions and the actuated joint inputs. It turns out that the algebra involved in the stability analysis may prove very complicated while the planar tensegrity mechanisms at hand are rather simple, thus making a complete classification intractable when standard computer algebra tools are used. To overcome this difficulty, more sophisticated tools are used, namely, discriminant varieties and cylindrical algebraic decomposition combined with Groebner bases. These tools allow classification of the number of stable solutions as a function of various parameters.

Next section 2 introduces the architecture of the family of planar tensegrity mechanisms under study. In section 3, the direct kinetostatic problem is first analyzed for different geometric and loading configurations. A deeper analysis is then conducted in section 4 with the help of appropriate computer algebra tools, where a full classification of the number of stable solutions is established. Finally, section 5 is devoted to the study of solution continuity, namely, in which conditions the mechanism at hand can move while always remaining in the same stable solution. Last section adds some discussions and concludes this work.

## 2    MECHANISM ARCHITECTURE AND EQUILIBRIUM CONDITION

The mechanism architecture under study is shown in Fig. 2. It is made of two crossed rigid rods $A_1A_3$ and $A_2A_4$ of lengths $L_1$ and $L_2$, respectively and three linear springs of stiffness $k$ connecting $A_1A_4$, $A_2A_3$ and $A_3A_4$. A reference frame is placed at the fixed point $A_1$ and the x-axis is along $A_1A_2$. Point $A_2$ can be translated along the x-axis with a prismatic actuator. This mechanism has three degrees of freedom. The first one, referred to as $\rho$, is controlled by the actuator to raise or lower the mechanism. The other two variables, $\theta_1$ and $\theta_2$, are generated by the compliant rotations of the two rods $A_1A_3$ and $A_2A_4$ about $A_1$ and $A_2$, respectively.

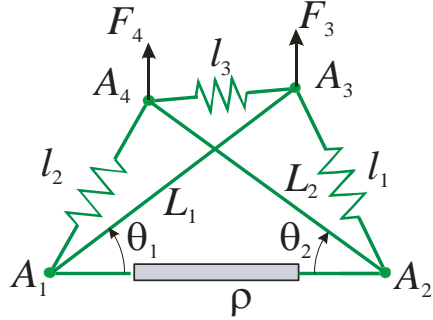

**Fig. 2: Planar tensegrity mechanism studied**

Two vertical forces $F_3$ and $F_4$ are applied at nodes $A_3$ and $A_4$, respectively. In this study, zero-free-length springs are considered. This is not a purely theoretical hypothesis since equivalent zero-free-lengths springs can be designed as shown for example in [8] and [12]. In the absence of friction and neglecting the compliance of the rods and of the prismatic joint, the potential energy $U$ of this mechanism can be written as follows:

$$U = \frac{k}{2} \sum_{i=1}^{3} l_i^2 - F_3 y_3 - F_4 y_4 \qquad (1)$$

where $y_3$ and $y_4$ are the ordinates of $A_3$ and $A_4$, respectively. Expressing the spring lengths $l_i$ as a function of the other geometric parameters using the cosine laws, $U$ takes the form below:

$$U = \frac{k}{2}\left(3\rho^2 - 4L_1\rho\cos(\theta_1) + 2(L_1^2 + L_2^2 + L_1L_2\cos(\theta_1 + \theta_2)) - 4L_2\rho\cos(\theta_2)\right) - F_3L_1\rho\sin(\theta_1) - F_4L_2\rho\sin(\theta_2) \qquad (2)$$

For the mechanism to be in equilibrium, the two derivatives of $U$ with respect to $\theta_1$ and $\theta_2$ must vanish simultaneously:

$$kL_1 \sin(\theta_2 + \theta_1) - 2k\rho\sin(\theta_2) + F_4 \cos(\theta_2) = 0 \qquad (3)$$

$$kL_2 \sin(\theta_2 + \theta_1) - 2k\rho\sin(\theta_1) + F_3 \cos(\theta_1) = 0 \qquad (4)$$

The solutions to the direct kinetostatic problem of the mechanism for a given input $\rho$, are obtained by solving Eqs (3) and (4) for $\theta_1$ and $\theta_2$, respectively.

## 3 SOLUTIONS TO THE DIRECT KINETOSTATIC PROBLEM

### 3.1 GENERAL CASE

The first step is to write equations (3) and (4) in polynomial form by using the tan-half-angle substitutions $t_1 = \tan(\theta_1/2)$ and $t_2 = \tan(\theta_2/2)$ :

$$F_3 t_1^2 t_2^2 + 2kL_2 t_1^2 t_2 + 2kL_2 t_2^2 t_1 + 4k\rho t_2^2 t_1 + F_3 t_1^2 - F_3 t_2^2 - 2kL_2 t_1 - 2kL_2 t_2 + 4k\rho t_1 - F_3 = 0 \tag{5}$$

$$F_4 t_1^2 t_2^2 + 2kL_1 t_1^2 t_2 + 2kL_1 t_2^2 t_1 + 4k\rho t_1^2 t_2 + F_4 t_1^2 - F_4 t_2^2 - 2kL_1 t_1 - 2kL_1 t_2 + 4k\rho t_2 - F_4 = 0 \tag{6}$$

Upon elimination of one of the variables (e.g. $t_2$) in the above two equations, a polynomial of degree six in $t_1$ is obtained after clearing the factor $(1+t_1^2)$. For each root, Eqs (5) and (6) can be combined to eliminate the terms of degree 2 and $t_2$ is then solved with a linear equation. Thus, the mechanism may have up to 6 solutions, of which some of them might be associated with instable equilibriums. Stable solutions are those solutions for which the 2×2 Hessian matrix $H$ is definite positive, namely, its leading principal minors $H(1,1)$ and $\det(H)$ are greater than zero. However, the general symbolic expressions of these minors are very large and stable solutions cannot be sorted out systematically at this stage.

We now inspect particular conditions that decrease the degree of the solution polynomial or lead to interesting, symbolically tractable cases.

### 3.2 SYMMETRIC DESIGN AND EQUAL FORCES

Here, a symmetric design $L_1 = L_2$ under symmetric loading $F_3 = F_4$ is examined. This situation was already studied by Arsenault [12], but under the assumption that the mechanism should always remain with its base and upper sides parallel, namely, all solutions should satisfy $y_3 = y_4$, or equivalently, $\theta_1 = \theta_2$. Accordingly, the direct kinetostatic problem was solved with only one equation, namely, the derivative of $U$ w.r.t. $y = y_3 = y_4$. A 4$^{th}$-degree polynomial equation was then obtained in $y = y_3 = y_4$ and solved for one particular set of parameters. In fact, it is not proven a priori that the solutions are always of the form $\theta_1 = \theta_2$. For the sake of exhaustiveness, the direct kinetostatic problem is solved here with $\theta_1$ not necessarily equal to $\theta_2$. Accordingly, the two equations (5) and (6) are used to solve the equilibrium conditions, in which $F_3$ and $L_2$ are replaced with $F_4$ and $L_2$, respectively. To get simpler expressions, the second equation is subtracted from the first one. The tan-half substitution is then applied in the resulting equation and the following new system is obtained:

$$(t_1 - t_2)(F_4(t_1 + t_2) + 2k\rho(1 - t_1 t_2)) = 0 \tag{7}$$

$$F_4(t_1^2 t_2^2 - 1) + 2L_1 k(t_1^2 t_2 + t_1 t_2^2) + 4k\rho(t_1 t_2 + t_2) + F_4(t_2^2 - t_1^2) - 2L_1 k(t_1 + t_2) = 0 \tag{8}$$

The first factor $(t_1 - t_2)$ in (7) confirms that solutions of the type $\theta_1=\theta_2$ do exist but the second factor indicates that solutions with distinct angles may also appear.

The solutions obtained from the first factor of (7) are calculated by substituting $t_2=t_1$ in eq (8), which yields up to 4 solutions.

The other solutions are obtained upon eliminating $t_1$ from the second factor of (7) and from (8), which leads to a polynomial of degree 2 in $t_2$:

$$4F_4 k^2 t_2^2 (L\rho + \rho^2) + 16k^3 \rho^3 t_2 + F_4^3(t_2^2 - 1) + 4F_4^2 k\rho t_2 + 4F_4 k^2 (L\rho - \rho^2) = 0 \tag{9}$$

Since $t_1$ can then been solved linearly using the second factor of (7), there are up to two solutions of the type $\theta_1 \neq \theta_2$. Moreover, the two solutions are of the form $(t_1, t_2)$ and $(t_2, t_1)$ since the same polynomial as (9) could have been obtained in $t_1$ by eliminating $t_2$ instead of $t_1$.

We now study the stability of the solutions. The leading principal minors $H(1,1)$ and $\det(H)$ of the Hessian matrix must be calculated and their sign must be positive for an equilibrium solution to be stable. The symbolic expression of $\det(H)$ for the solutions $\theta_1 \neq \theta_2$ is surprisingly simple and the sign analysis can be handled easily. Indeed, solving the second factor of (7) for $t_2$ and replacing the solution in $\det(H)$ leads to the following factored expression:

$$\det(H) = -\frac{4(t_2^2+1)^2 (4k^2\rho^2 + F_4^2)^2 (-k\rho t_2^2 + F_4 t_2 + k\rho)^2}{(-2k\rho t_2 + F_4)^4} \tag{10}$$

It can be observed that $\det(H)$ is always negative. Thus, the two equilibrium solutions of the type $\theta_1 \neq \theta_2$ are always unstable. On the other hand, the sign study of $\det(H)$ and $H(1,1)$ for the solutions of the type $\theta_1=\theta_2$ is not straightforward and requires tedious analyses. Stability analysis of those solutions is left to section 4 with the use of more appropriate tools.

### 3.2 NO EXTERNAL LOADING ($F_3=F_4=0$)

When $F_3=F_4=0$, Eqs (3) and (4) become:

$$L_1 \sin(\theta_2 + \theta_1) - 2\rho \sin(\theta_2) = 0 \tag{11}$$

$$L_2 \sin(\theta_2 + \theta_1) - 2\rho \sin(\theta_1) = 0 \tag{12}$$

It is apparent that $\theta_i = 0$ or $\pi$, i=1, 2 are solutions to the above system, which define four singular configurations. In such configurations, the mechanism is fully flat and cannot resist any vertical forces. There are two more solutions of the form:

$$(\theta_1 = \arctan(Q/\rho L_1), \theta_2 = \arctan(Q/\rho L_2)), \ (\theta_1 = -\arctan(Q/\rho L_1), \theta_2 = -\arctan(Q/\rho L_2)) \tag{13}$$

where $Q = \left((L_1+L_2)^2 - 4\rho^2\right)\left((L_1-L_2)^2 - 4\rho^2\right)$. On the other hand, when the coordinates of $A_3$ and $A_4$ are calculated with the above solutions, one obtains $y_3 = y_4$ and $x_3 - x_4 = \rho$. This means that the mechanism remains always in a parallelogram configuration, even when $L_1 \neq L_2$.

It can be shown easily by reporting $\theta_{1,2} = 0$ or $\pi$ into det($H$) and $H(1,1)$, that three solutions of the four flat ones are always unstable.

## 4 Solutions classification

We know from previous section that the solutions are obtained upon solving a polynomial of degree six. In some particular cases, we were able to identify stable solutions. The goal of this section is to classify, in a systematic way, the number of stable equilibrium solutions as a function of the geometric and physical parameters of the planar tensegrity mechanisms. The algebraic problem at hand relies on solving a polynomial parametric system of the form:

$$E = \{\mathbf{v} \in \mathbb{R}^n, p_1(\mathbf{v}) = 0, \ldots, p_m(\mathbf{v}) = 0, q_1(\mathbf{v}) > 0, \ldots, q_l(\mathbf{v}) > 0\} \tag{14}$$

Such systems can be solved in several ways. Discriminant varieties [13, 15, 16] and cylindrical algebraic decomposition [14, 15, 16] are used here. They provide a formal decomposition of the parameter space through an algebraic variety that is known exactly. These tools have already been applied successfully for the classification of the singularities of serial and parallel manipulators [15, 16], a class of problems that can be formulated as in (14). Shortly speaking, discriminant varieties generate a set of separating hyper-surfaces in the parameters space of the parametric system at hand, such that the number of solutions in each resulting connected component or cell is known and constant. The discriminant varieties can be computed with known tools like Groebner bases using the Maple sub-package *RootFinding[parametric]*. Once the discriminant varieties are obtained, an open cylindrical algebraic decomposition is computed to provide a full description of all the cells: the number of solutions in each cell is determined by solving the polynomial system for one arbitrary point in each cell. Finally, adjacent cells with the same number of solutions are merged.

The equilibrium solutions depend on three geometric parameters (the rod lengths $L_1$ and $L_2$ and the input variable $\rho$) and two physical parameters (the spring stiffness k and the forces $F_3$ and $F_4$). Without loss of generality, however, the lengths parameters can be normalized with $L_1$, while $F_3$ and $F_4$ can be normalized with $k$. Finally the system at hand depends on four independent parameters only. In what follows, $L_1$ and $k$ are fixed to 1 and 100, respectively, without loss of generality.

## 4.1 NO EXTERNAL LOADING

In this section, the unloaded situation $F_3=F_4=0$ is studied. We were able to show in the preceding section that three of the four flat solutions were unstable but no general information could be obtained about the two non-flat solutions. Since $F_3=F_4=0$, the parameter space is a plane defined by $L_2$ and $\rho$. Calculating the discriminant varieties and the cylindrical algebraic decomposition for this case shows that there exists one region in the parameter plane where the mechanisms have two stable solutions. Outside this region, only one stable solution exists. Figure 3 (left) shows a representation of the parameter plane obtained for $L_2$ and $\rho$ in [0, 4]. The 2-solution region is shown in red. The three separating lines that bound the regions are defined by $2\rho-L_2-1=0$, $2\rho-L_2+1=0$ and $2\rho+L_2-1=0$, respectively. Here it can be verified with simple geometric arguments that these boundaries define the limit conditions for the mechanisms to become fully flat. In the three 1-solution regions, the mechanisms have one stable fully flat solution. In those regions including their boundaries, the mechanisms are thus in singular configurations. They can move along a horizontal direction but they cannot resist any vertical force. Such solutions could be used to store the mechanisms when there are not in use. In the 2-solution regions, there are two stable non-flat solutions, one being the mirrored image of the other as shown in Fig. 3 (right). Moreover, as already shown in section 3.1, the mechanisms take the shape of a parallelogram even when $L_1 \neq L_2$. The 2-solution region is of constant width equal to 1 when $L_2>1$, while it decreases linearly when $L_2<1$. In view of the optimal design, it is thus interesting to note that the largest operation range $\Delta\rho = 1$ is obtained for $L_2>1$.

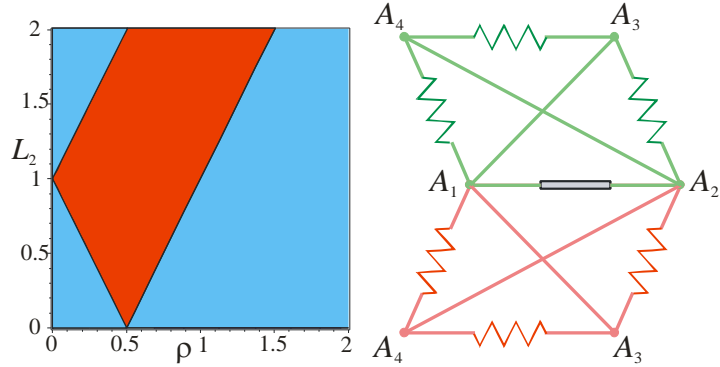

**Fig. 3: Unloading case: parameter plane (left) and the two mirrored stable solutions for $\rho=1$, $L_2=3/2$ (right)**

## 4.2 FULLY SYMMETRIC CASE

The mechanisms have now a symmetric design ($L_1=L_2$) and symmetric loading conditions ($F_3=F_4$). It was shown in the preceding section that the two solutions $\theta_1 \neq \theta_2$ were always unstable but no stability information could be brought for the four solutions of the type $\theta_1 = \theta_2$. The parameter space is now the plane ($\rho$, $F_4$), since $L_1=L_2=1$. The computed discriminant varieties and cylindrical algebraic decomposition provides the parameter plane shown in fig. 4a (left) for $0<\rho\varepsilon 2$, $-10\varepsilon\ F_4\varepsilon 0$ (pushing forces) and in fig. 4b for $0<\rho\varepsilon 2$, $0\varepsilon\ F_4\varepsilon 10$ (pulling forces).

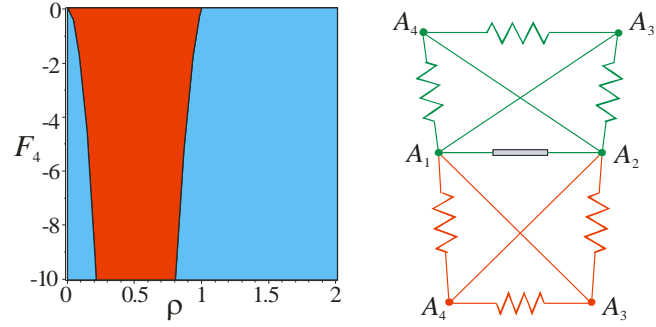

**Fig. 4a: Symmetric case: parameter plane under pushing forces for $0<\rho\varepsilon 2$, $-10\varepsilon\ F_4\varepsilon 0$ (left) and two stable solutions for $\rho=3/4$, $F_4=-10$ (right)**

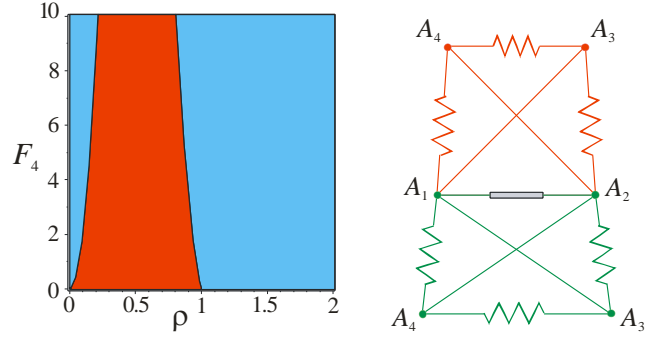

**Fig. 4b: Symmetric case: parameter plane under pulling forces for $0<\rho\varepsilon 2$, $0\varepsilon F_4\varepsilon 10$ (left) and two stable solutions for $\rho=3/4$, $F_4=10$ (right)**

There exists a region with two stable solutions and two regions with one stable solution, filled in red and blue, respectively. Since we know that the solutions corresponding to $\theta_1 \neq \theta_2$ are always unstable, this means that two or three of the four solutions of the type $\theta_1 = \theta_2$ are unstable. Figure 4a, right (resp. 4b, right) illustrates two stable solutions for $\rho = 3/4$ and $F_4 = -10$ (resp. $\rho = 3/4$ and $F_4 = 10$). Contrary to the unloaded case, those two solutions are not symmetric.

The boundaries between the 2- solution and the 1-solution regions are two curves of degree 6 defined by:

$$F_4^6 + 12\times 10^4 F_4^4 \rho^2 + 48\times 10^8 F_4^2 \rho^4 + 64\times 10^{12} \rho^6 - 16\times 10^8 F_4^2 \rho^2 = 0 \qquad (15)$$

$$F_4^6 + 12\times 10^4 F_4^4 \rho^2 + 48\times 10^8 F_4^2 \rho^4 + 64\times 10^{12} \rho^6 - 12\times 10^4 F_4^4 + 336\times 10^8 F_4^2 \rho^2 \\ -192\times 10^{12}\rho^4 + 48\times 10^8 F_4^2 + 192\times 10^{12}\rho^2 - 64\times 10^{12} = 0 \qquad (16)$$

In the two 1-solution regions, it can be shown that the mechanisms operate always with $y_3$ and $y_4$ negative under pushing forces. Figure 4c shows stable solutions selected in each of the 1-solution regions. Assuming that a mechanism is controlled to start in a configuration with $y_3$ and $y_4$ positive, its operation range $\Delta\rho$ for a given pushing $F_4$ is thus determined by the 2-solution region only. The operation range decreases when the magnitude of external force increases. It can be verified that for $F_4=0$, $\Delta\rho$ reaches its maximal value, equal to 1, in accordance with the result in section 3.1. In the presence of pulling forces ($F_4>0$), $y_3$ and $y_4$ are positive in the two one-solution regions. Figure 4d shows stable solutions chosen in each of the two 1-solution regions.

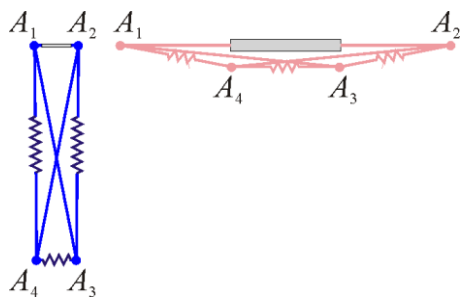

**Fig. 4c: Stable solution in each of the two 1-solution regions under a pushing force $F_4=-10$ (left: $\rho=2/10$, right: $\rho=3/2$)**

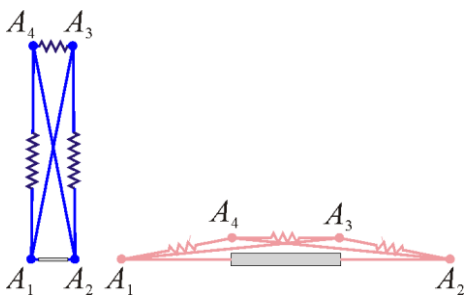

**Fig. 4d: Stable solution in each of the two 1-solution regions under a pulling force $F_4=10$ (left: $\rho=2/10$, right: $\rho=3/2$,)**

## 4.3 GENERAL CASE

The parameter space is now 4-dimensional since it is defined by ($\rho$, $L_2$, $F_3$, $F_4$). One parameter is assigned in order to have a parameter space of dimension 3. First, the design parameter $L_2$ is fixed. Figure 5a (resp. 5b, 5c) shows the obtained partition of the parameter space ($\rho$, $F_3$, $F_4$) when $L_2=3/2$ (resp. 1, 1/2) for $0<\rho\leq 2$ and $0\leq F_3 \leq 10$, $0\leq F_4 \leq 10$. The boundaries are three surfaces of degree 12 in $\rho^2$ and in $F_3$, $F_4$, defined by equations containing not less than 436 terms. Note that the regions have been also plotted on separate figures to show better the separating surfaces.

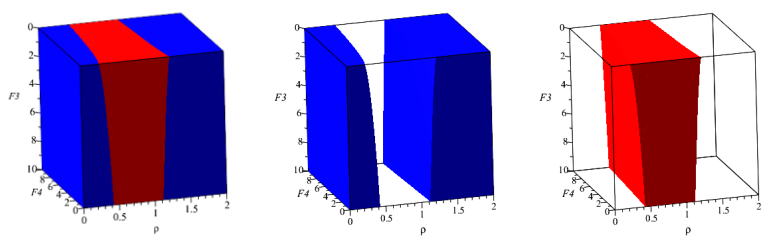

**Fig. 5a: Partition of the parameter space when $L_2=3/2$ for $0<\rho\leq 2$ and $0\leq F_3 \leq 10$, $0\leq F_4 \leq 10$**

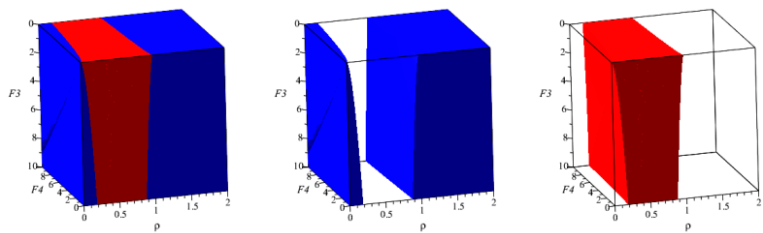

**Fig. 5b: Partition of the parameter space when $L_2=1$ for $0<\rho\leq 2$ and $0\leq F_3 \leq 10$, $0\leq F_4 \leq 10$**

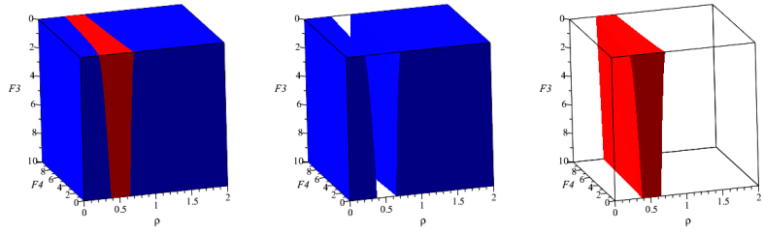

**Fig. 5c: Partition of the parameter space when $L_2=1/2$ for $0<\rho\leq 2$ and $0\leq F_3 \leq 10$, $0\leq F_4 \leq 10$**

These plots show that the regions with two stable solutions have a regular shape and are of reasonable size under quite small force magnitudes. It is apparent that when $L_2$ is increased, the size of those regions is also increased. It is interesting to see how the size of the 2-solution regions varies when much higher forces are applied. Figure 6 shows how the 2-solution region shrinks and even disappears when the force magnitudes reach certain limits. Since the discriminant varieties give an algebraic description of the separating surfaces, it is possible to determine exactly the singular points of these surfaces and in particular the coordinates of transition point C.

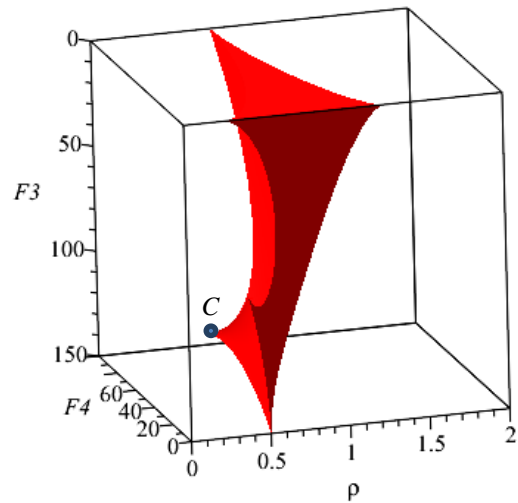

**Fig. 6: The 2-solution region for large ranges of force magnitudes. Point C defines the limit force magnitudes for two stable solutions to exist.**

Two parameters are now assigned in order to have a parameter space of dimension 2. Accordingly, the discriminant varieties and cylindrical algebraic decomposition are computed for $F_3=F_4=-10$. Figure 7 (left) shows the obtained partition of the parameter plane $(\rho, L_2)$ for $0<\rho\leq 2$ and $0\leq L_2\leq 2$. It looks similar to the unloaded case but the boundaries here are three curves of degree 12 in $\rho$ and in $L_2$ and their equations are too large to be displayed in this paper. There are two stable solutions in the red region and only one in the blue region. Figure 7 (right) shows two stable solutions for $L_2=3/2$ and $\rho=7/10$.

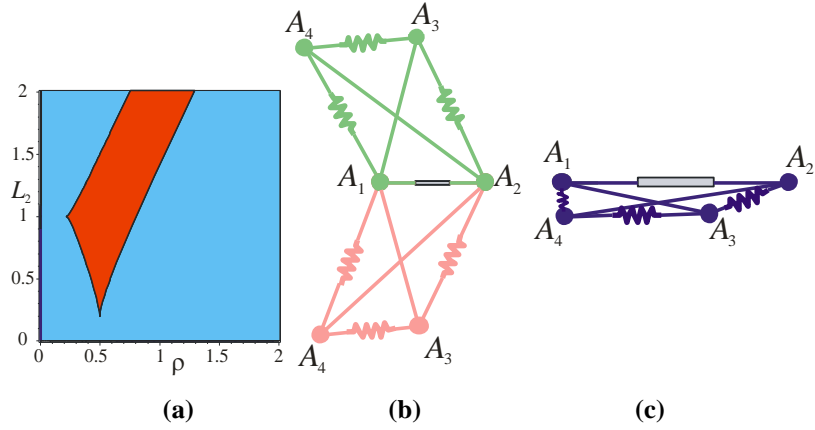

**Fig. 7:** Slice of the parameter space for $F_3=F_4=-10$ (a), the two stable solutions when $L_2=3/2$ and $\rho=7/10$ (b) and the stable solution when $L_2=3/2$ and $\rho=3/2$ (c)

Like in the preceding case, the operation range is determined by the 2-solution region if the mechanism starts with $y_3>0$ and $y_4>0$. The operation range reaches its maximal width for $L_2=1$, which is the fully symmetric case: it can be verified that this range is exactly the same as the one calculated in the above section for $F_3=F_4=-10$. Starting from $L_2=1$, the operation range then decreases slowly when $L_2$ increases but the decrease is much more significant when $L_2$ decreases.

## 5 Solution continuity

Section 4 above shows that one or two stable solutions may exist for a given value of the actuator parameter $\rho$. If a mechanism is actuated to follow a prescribed path, it is necessary to verify that the starting solution remains available during the whole trajectory to avoid any undesirable jump to another solution during motion. In this section, the solution continuity is analyzed. The case with no external forces is clear since the 1-solution regions are associated with only singular configuration that cannot be used for practical motions and the mechanisms must be operated by starting and ending any trajectory in the 2-solution region only. In the fully symmetric case, the 2-solution region is surrounded with two 1-solution regions. Figure 8a shows the evolution of the solutions for a mechanism defined by $L_1=L_2=1$ and under pulling forces $F_3=F_4=5$ when the actuator parameter $\rho$ varies from 0 to 2. Both stable solutions (in blue) and unstable solutions (in red) are depicted. The plots show two interesting critical points $C_1$ and $C_2$ at the extremities of the lower blue curve (the negative solutions). At $C_1$, the solution turns unstable when $\rho$ is decreased. Moreover, $C_1$ is a branching point where two distinct unstable solution branches become available. In practice, the mechanism will jump to the other stable solution (the one with a positive $\theta_1$) if $\rho$ is decreased. At $C_2$, the stable solution curve merges with an unstable solution curve at a fold point. If $\rho$ is increased, the mechanism will jump again to a positive stable solution. At both critical points, a discontinuous behavior of the mechanism is experienced under a continuous change in the parameters, which is known as a *catastrophe*, a phenomenon often observed in mechanisms with springs [12, 17]. Figure 8b shows that when $L_2$ is slightly increased, the first critical point $C_1$ is transformed into a fold point.

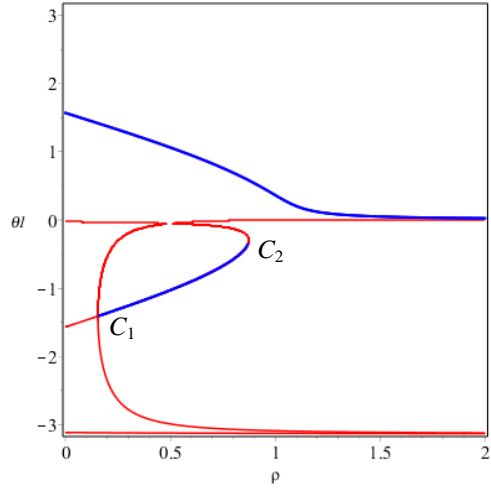

**Fig. 8a: Solution continuity with $L_1=L_2=1$, $F_3=F_4=5$ (stable solutions in blue)**

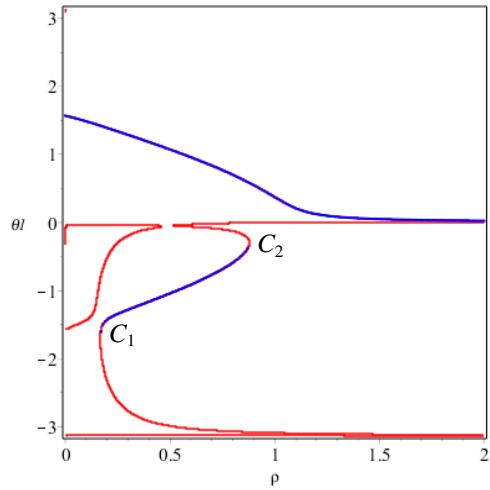

**Fig. 8b: Solution continuity with $L_1= 1$, $L_2=1.05$, $F_3=F_4=5$ (stable solutions displayed in blue)**

## 6     DISCUSSION

External forces have been supposed to be vertical in this work. When horizontal force components $F_{3x}$ and $F_{4x}$ are added, it can be shown that this does not change the global nature of the algebraic equations and of the results. Indeed the only changes are the additional term $2F_{3x}$ (resp. $2F_{4x}$) appearing in the coefficients of $t_2^2 t_1$ and of $t_1$ in Eq. (5) (resp. of $t_1^2 t_2$ and of $t_2$ in Eq. (6)). Globally one comes up with a system yielding six solutions, of which one or two are stable like. The gravity has been omitted in this work. Again, taking into account the rod loads does not modify the nature of the problem. It just adds additional terms in the equations that result in slightly modified equilibrium solutions. The three springs were assumed similar but taking different stiffness's does not change the nature of equations and of the general results as well. In fact, playing with additional force components or distinct stiffness values increases the dimension of the parameter spaces. This is not problematic for the calculation of the discriminant varieties but the computation time of the cylindrical algebraic decomposition increases and graphical representations are more challenging. Using springs with non-zero free lengths make the equations much more complex and might render symbolic calculations intractable, as shown in [11] and already observed in former works [8, 9]. Practically, however, equivalent zero-free lengths springs can be designed without much effort. Finally, interferences between the two crossed rods can be easily avoided by assembling them in two different layers, as shown in [12] for example.

## 7 CONCLUSIONS

The goal of this paper was to investigate in depth the direct kinetostatic solutions of a family of planar tensegrity mechanisms composed of a telescopic base, two crossed rods and three springs. It was shown in this paper that the problem can be treated using suitable computer algebra tools. We have shown that a univariate polynomial of degree 6 must be solved in the general case, resulting in one to two stable solutions. In the unloaded case, there are always two stable symmetric solutions for a range of the input prismatic joint which is of constant width and whose limits vary with the rod lengths. Moreover, the mechanisms remain always in a parallelogram configuration even when their two rod lengths are different. The mechanisms can also reach one flat stable solution, which is singular. Such a stable flat solution might be of interest to store the mechanism when it is not used. When the two external forces and the two rod lengths are equal, there are still 6 solutions, including 4 unstable, non-symmetric solutions.

We have used discriminant varieties and cylindrical algebraic decomposition to study the evolution of the number of solutions as function of various parameters. Those tools have already proved efficient to treat several difficult kinematic issues arising in serial and parallel manipulators. Discriminant varieties and cylindrical algebraic decomposition generate a set of separating curves or surfaces in the parameter space, which are fully defined with algebraic equations.

The next step of this work is to consider and design an assembly of proper planar tensegrity modules stacked in series and suitable actuation to get a first planar model of a bird neck. There are many different choices in the design of the elementary tensegrity module by playing with the choice of the actuated link(s) and the number of springs [12]. It is possible to actuate a second rod, namely, one of the crossed ones. The classification analysis conducted in this paper can be applied directly to those mechanisms. It is also possible to select a mechanism with only two springs along the two sides and all other segments made with fixed-length rods. Such a mechanism has only one degree of freedom in total (a rotation about the intersection point of the two crossing rods) and can be driven with cables [18]. Figure 9 illustrates a possible assembly of three such mechanisms with two driving cables thread through the spring attachment points as described in [18]. Depending on the number of parameters to be controlled (position only or position and orientation of the terminal link, stiffness, configuration of the whole serial assembly), the arrangement and actuation of the cables have to be properly chosen.

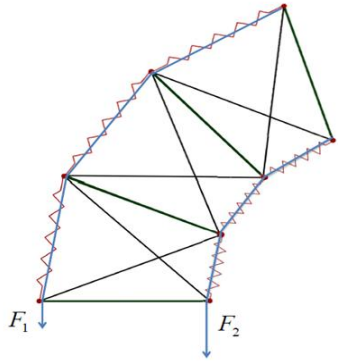

**Fig. 9: An assembly of 2-spring tensegrity mechanisms.**

**Acknowledgments** This work was funded by the French ANR project "AVINECK: an arm for the bird", ANR-16-CE33-0025-02, 2017-2020.